\documentclass{article}
\usepackage{ijcai18}

\usepackage{times}
\usepackage{xcolor}
\usepackage{soul}
\usepackage[utf8]{inputenc}
\usepackage[small]{caption}

\usepackage{booktabs} % For formal tables
\usepackage{algorithm}  
\usepackage{algorithmicx}  
\usepackage{algpseudocode}  
\usepackage{amsmath}  
\usepackage{amssymb}  
\usepackage{graphicx}
\usepackage{subfigure}
\usepackage{verbatim}
\usepackage{array}
\usepackage{bm}
\usepackage[colorinlistoftodos]{todonotes}
\usepackage{threeparttable}

\newcommand{\ty}[1]{\hat{y}_{#1}}

\newcommand{\sF}[1]{f(x_{#1}; \theta)}

\newcommand{\SampleSet}{\mathbb{S}}

\newcommand{\SoftAUC}{SAUC}
\newcommand{\EmpSoftAUC}{\widehat{\SoftAUC}}
\newcommand{\EmpSoftAUCExp}{\widehat{\SoftAUC}(\theta; \SampleSet)}

\newcommand{\SwapIndic}[5]{#1 ( #2{#3} #5 #2{#4} )}

\newcommand{\SwapIndicSoftY}[2]{\SwapIndic{\sigma}{\ty}{#1}{#2}{-}}
\newcommand{\SwapIndicSoftF}[2]{\SwapIndic{\sigma}{\sF}{#1}{#2}{-}}

\newcommand{\SwapGainSym}[2]{(y_{#1} - y_{#2})}

\title{Learning Theory and Algorithms for Revenue Management in Sponsored Search}
\author{
Lulu Wang, 
Huahui Liu, 
Guanhao Chen, 
Shaola Ren,
Xiaonan Meng,
Yi Hu
\\ 
Alibaba Group, Hangzhou, 310052, China \\
\{sengyun.wll, huahui.lhh, lea.cgh, shaola.rs, xiaonan.mengxn, erwin.huy\}@alibaba-inc.com
}
\begin{document}
\maketitle

\begin{abstract}
Online advertisement is the main source of revenue for Internet business. 
Advertisers are typically ranked according to a
score that takes into account their bids and potential click-through
rates (eCTR). Generally, the likelihood that a user clicks on an ad is often modeled by optimizing for the click through rates rather than the performance of the auction in which the click through rates will be used. This paper attempts to eliminate this disconnection by proposing loss functions for click modeling that are based on final auction performance. In this paper, we address two feasible metrics ($AUC^R$ and $SAUC$) to evaluate the online RPM (revenue per mille) directly rather than the CTR. And then, we design an explicit ranking function by incorporating the calibration factor and price-squashed factor to maximize the revenue. Given the power of deep networks, we also explore an implicit optimal ranking function with deep model. Lastly, various experiments with two real world datasets are presented. In particular, our proposed methods perform better than the state-of-the-art methods with regard to the revenue of the platform.
\end{abstract}

\section{Introduction}
Online advertisement (AD) is the main source of revenue for Internet business. After years of evolution, the mechanism of AD has changed from the pre-allocated style to keyword-based matching of \textit{Sponsored (or paid) search}. Sponsored search such as Google AdWords and Bing's Paid Search, is search advertising that shows ads alongside algorithmic search results on search engine results pages (SERPs). Sponsored search has evolved to satisfy users' need for relevant search results and advertisers' desire for qualified traffic to their websites. And it is now considered to be among the most effective marketing vehicles available.

Generally, the problem of revenue optimization is typically framed as a question of finding the revenue-maximizing incentive compatible auction \cite{introduction1,introduction2,introduction3,introduction4,introduction5}. Advertisers are typically ranked according to a score that takes into account their bids and the potential click-through
rates. The revenue space in sponsored search can be divided into two parts: click-through rates prediction (eCTR) and optimal bidding. The prediction tries to estimate the user's behavior as accurately as possible and it occupies an important position in the advertising system. Bidding tries to find a optimal price for each impression and it is closely related to the ROI (\textit{Return on Investment}) of advertisers and revenue of advertising platforms. There are also previous works focused on the relationship between eCTR and bid and attempted to maximize the revenue through this way \cite{lahaie-1,lahaie-2}. A number of variants (with parameters) have been used in practice. And almost all the works explore this space of parametrized mechanisms, searching for the optimal designs. However, most of the methods and their variants stay on theoretical analysis level. The advertising system, including our platforms, tunes the parameters in a sandbox environment until the performance converges. The efficiency of this approach is extremely low, and usually we cannot reach the optimal point.

\begin{comment}
The key to the problem is that we lack an offline indicator to evaluate the online revenue directly. Given that the existing offline metrics (AUC, LOGLOSS and other metrics) do not measure the online performance well, in this paper, we propose two offline evaluation metrics i.e., $AUC^R$ and $SAUC$ to simulate the online revenue directly. Lots of online experiments show the effectiveness of the metrics as indicators for the RPM of the platform. Then we turn the problem of revenue optimization into the problem of maximization of the offline evaluation metrics. 

Based on the previous research results, we introduce two parameters,
${\vec\alpha}=\mathbf{[\alpha_1,...,\alpha_n]}^\mathrm{T}$ and
${\vec\beta}=\mathbf{[\beta_1,...,\beta_n]}^\mathrm{T}$, 
to depict the ranking function and give a closed-form function of the topic. In our work, the parameter $\vec\alpha$ is used for calibration of CTRs and $\vec\beta$ is used for the optimization of revenue. Then two efficient algorithms, SOR (\textit{Two-stage optimization for revenue)} and GOR (\textit{Global optimization for revenue)}, are given to learn all the parameters from the historical data.
\end{comment}

 This paper mainly focuses on the problem of revenue management. Our main contributions are summarized as follows:
\begin{enumerate}
\item We propose loss functions for click modeling that are based on final auction performance.  From the view of the loss function, we address two metrics ($AUC^R$ and $SAUC$) to indicate the online RPM. To our knowledge, this is the first paper in open literature that tries to evaluate the online RPM rather than the CTR directly.

\item We explore the implicit and explicit ranking functions to maximize the RPM in sponsored search. Experiments and discussions on two real-world advertising platforms show consistent improvement over existing methods.

\end{enumerate}
\begin{comment}
The remaining parts of this paper are organized as follows. In the next section, we start by formally stating the problem to be addressed. We briefly introduce the related works about revenue management problem and address a general formulation for revenue management problem. In section 3, we will cover the methodology in more detail and show how it can be implemented. In section 4, we show our experiments for benchmarking the proposed methods and make a detailed discussion for the results. Finally, we summarize our findings and suggest best practice guidelines based on our analysis and lessons from real world experiences on online advertising system. 
\end{comment}

\section{Preliminaries}

\subsection{Related Work}
The target application of our study is online advertising. Some of the problem issues discussed in this study might be specific to the domain. In this section, we briefly introduce previous works related to the revenue management problem.

\subsubsection{Offline Evaluation Metric}
We studied papers from the proceedings of the International World Wide Web Conference (WWW), the ACM International conference on Web Search and Data Mining Conference (WSDM), the International Joint Conference on Artificial Intelligence (IJCAI), and the ACM SIGKDD International Conference on Knowledge Discovery and Data Mining Conference (SIGKDD) in years 2016 and 2017 in the area of algorithmic search and online advertising. 
We  manually categorized the topic areas of the papers and the evaluation metrics they used, and summarizes that AUC \textit{(the Area Under the Receiver Operating Characteristic Curve)} is the most widely used evaluation metric\footnote{As far as we know, most productive advertising platforms also apply AUC as the offline evaluation metric.}. 

\begin{comment}

One of the important topics in sponsored search is to find a proper offline evaluation metric to indicate the online performance (\textit{RPM, revenue per mille}). In practice, the likelihood that a user clicks on an ad is often modeled using traditional machine learning techniques and, in particular, traditional machine learning loss functions. 
Then, we will use an offline evaluation metric to indicate the online performance so that we can expect results without online experiments.

For a typical machine learning problem, training and evaluation (or test) samples are selected randomly from the population the model needs to be built for, and predictive models are built on the training samples. Then the learned models are applied on the evaluation data, and the qualities of the models are measured using selected evaluation metrics. The evaluation metric is so significant that we can expect results without online experiments. 
Traditionally, the advertising system use AUC \textit{(the Area Under the Receiver Operating Characteristic Curve)} or LOGLOSS as the offline evaluation metrics. 
\end{comment}

In fact, AUC is potentially suboptimal because the goal, in sponsored search, is not to optimize click action but instead to optimize performance of the auction in which the click through rates will be used. In view of the bidding, we often experience much more discrepancy between the offline (AUC for eCTR) and online performance (RPM) \cite{offline-matrix1}. In this paper, we attempt to eliminate the disconnect by proposing loss functions for click modeling that are based on final auction performance. The experimental results show that the proposed evaluation metrics are highly promising.

\subsubsection{Ranking Functions and Auction Mechanism}
There are several works investigating on how to learn a reasonable ranking function to maximize the revenue \cite{optimize-1,optimize-2,optimize-3,optimize-4,optimize-5}. The revenue management in sponsored search is closely related to the auction mechanism. As the stakes have grown, the auction mechanism in sponsored search has seen several revisions over the years to improve efficiency and revenue. When first introduced by GoTo in 1998, ads were ranked purely by bid. In \textit{cost-per-click (CPC)} advertising system, a parametrized family of ranking rules that order ads according to rank score is shared by every major search engine now. 
\begin{equation}rankscore=eCTR*b\end{equation}
where $b$ is a bid amount, $eCTR$ is the estimated position-unbiased CTR \cite{rank-function2}. Rank score is the estimated CTR weighted by a cost per click bid.
In last decade, there has been an intense research activity in the study of the CTR prediction  \cite{lr-ctr3,tree-ctr1,deep-ctr3,deep-ctr5}.

Under the assumption that CTRs are measured exactly, it is simple to verify that ranking ads in order of eCTR times bid is economically efficient. However, it is hard to measure the CTRs exactly no matter how many efforts you put in \cite{offline-matrix1,bias-2}. To address this, previous work mainly focused on two points:
\begin{enumerate}
\item \textbf{Calibration methods} In \cite{lr-ctr1}, they use a calibration layer to match predicted CTRs to observed click-through rates. And they improve the calibration by applying a correction functions $\tau_d (p)=\gamma{p}^k$ where $p$ is the predicted CTR and $d$ indicates a partition of the training data. They use isotonic regression on aggregated data to learn $\gamma$ and $k$.

\item \textbf{Price-squashed factor} Lahaie and Pennock propose a parametrized family of ranking rules that order ads according to scores of the form
\begin{equation}rankscore=eCTR^\beta*b\label{fun:2}\end{equation}
where $\beta$ is a parameter, called \textit{price-squashed factor} or \textit{click investment power} \cite{lahaie-1}. 
If $\beta > 1$,  the auction prefers ads with higher estimated CTRs, otherwise, ads with higher bids. Further, Lahaie show that, in the presence of CTR uncertainty, using $\beta$ less than 1 can be justified on efficiency grounds\cite{lahaie-2}. 
\end{enumerate}

In this paper, we incorporate both of the two factors to develop an explicit ranking function for revenue management. Further, given the power of deep network, we try to learn a reasonable ranking function with deep model.

\subsection{Problem Formulation}
Suppose that the training data is given as lists of feature vectors (refer in particular to eCTR and bid here) and their corresponding lists of labels ($revenue=click*bid$, $click\in{[0,1]}$), $S=\{(x_i,y_i)\}_{i=1}^m$. We are to learn a ranking model $f(x)$ defined on object (feature vector) $x$. Given a new list of objects (feature vectors) $\textbf{x}$, the learned ranking model can assign a score to each $x$, $x\in\textbf{x}$. And then sort the objects based on the scores to generate a ranking list (permutation) $\pi$. The evaluation is conducted at the list level, specifically, a evaluation measure $E(\pi,y)$ is utilized. 

\textbf{\textit{Definition}} The revenue management is to optimize the ranking accuracy in terms of a performance measure on the training data:
\begin{equation}
\begin{aligned}
\label{fun:3}
f^* &= \arg \max_{f\in\digamma} Revenue \\
&=\arg \max_{f\in\digamma}{\sum_{i=1}^{m}E(\pi(eCTR_i,bid_i,f),y_i)} 
\end{aligned}
\end{equation}
where $\digamma$ is the set of possible ranking functions, i.e., \textbf{function (\ref{fun:2})}, m denotes the number of samples and $E(\pi,y)$ means the offline evaluation metrics. 

From the definition, we can illustrate that the key points of the revenue management are: 1) an optimal ranking functions; 2) a metric to indicate the online performance . This paper will focus on the two aspects.

\section{Methodology}
\subsection{Offline Evaluation Metric} 

\subsubsection{AUC and Loss Function}
AUC is defined as the expectation of ranking a randomly chosen positive sample above a randomly chosen negative one. It is a popular metric for ranking performance evaluation and is extensively used in regression problems with binary labeled samples (e.g. CTR prediction). 
We can further understand AUC in the view of loss function. The loss in organic search (model the click action only) is shown in \textbf{Table 1}. In the table, $clk_y$ means an ad is clicked and $clk_n$ means not. The pair ($clk_n$, $clk_y$) means that a negative sample ranks above a positive one, and we miss a click.

\begin{table}
\centering
\caption{Loss in \textit{organic search} ranking}
\label{tab:1}
\begin{tabular}{p{3cm}<{\centering}|p{3cm}<{\centering}} \hline
\textbf{Group}&\textbf{Loss}\\ \hline \hline 
$(clk_{y},clk_{y})$   &        0\\ \hline
$(clk_{y},clk_{n})$   &         0\\ \hline
$(clk_{n},clk_{y})$   &         1\\ \hline
$(clk_{n},clk_{n})$   &          0\\
\hline\end{tabular}
\end{table}

Let $x_i$ be the features of sample $i$. Let $y_i$ be the label of sample $i$ with $1$ and $0$ for positive and negative samples respectively.
Let $\widehat{y_i}=f(x_i;\theta)$ be the predicted ranking score of sample $i$. To generalize the above table, in the view of loss function, we can address a formal definition of AUC. 
\begin{equation}
\begin{aligned}
\label{func:4}
  AUC =\frac{1}{\lambda} \mathbb{E}[  & \mathbb{I}{(\widehat{y_i}>\widehat{y_j})}max(0,y_i-y_j) + \\
                      & \mathbb{I}{(\widehat{y_j}>\widehat{y_i})}max(0, y_j-y_i) ]
\end{aligned}
\end{equation}
where $\mathbb{I}$ is the indicator function, $\mathbb{E}$ means the expectation of an event and $\lambda=\frac{M*N}{(M+N)^2}$ (for rigor and can be ignored). M and N is the number of positive and negative samples respectively. 
\begin{comment}
\mathbb{I}{(\widehat{y_i}>\widehat{y_j})}\mathbb{I}{(y_i=1, y_j=0)} +\\
                      & \mathbb{I}{(\widehat{y_j}>\widehat{y_i})}\mathbb{I}{(y_j=1,y_i=0)}] \\ 
\end{comment}

Due to its discrete nature, AUC is neither applicable to real value labeled samples (e.g. RPM ranking) nor could be optimized directly. However, the formalization of AUC sheds light on two straight forward ideas.
\begin{enumerate}
\item \textbf{Real-Valued AUC} ($AUC^R$) :
  the $y_i$ is not necessarily being binary and
  the AUC could be naturally extended for problems with real value labeled samples.
\item \textbf{Soft AUC} ($SAUC$) :
  by replacing the discrete indicator function with its continuous approximation(e.g. sigmoid),
  the AUC itself could be optimized with gradient based methods.
\end{enumerate}

\subsubsection{Real-Valued AUC and Loss Function} Different from the rank order by pure eCTRs, the problem involves the bid factor in sponsored search (\textbf{Table 2}). In the table, $b_i$  is the bidding for $sample_i$ and $clk_y$ means the ad is clicked, (($clk_n,b_i$), ($clk_y,b_j$)) means that a negative $sample_i$ ranks a above positive $sample_j$, and we lost $\max(b_j*1-b_i*0,0)$ revenue.

\begin{table}[!htb]
\centering
\caption{Loss in \textit{sponsored search} ranking}
\label{tab:2}
\begin{tabular}{c|c} \hline
\textbf{Group}&\textbf{Loss}\\ \hline \hline 
$((clk_{y},b_1),(clk_{y},b_2))$ & $\max(b_2*1-b_1*1,0)$\\ \hline
$((clk_{y},b_1),(clk_{n},b_2))$ & $\max(b_2*0-b_1*1,0)$\\ \hline
$((clk_{n},b_1),(clk_{y},b_2))$ & $\max(b_2*1-b_1*0,0)$\\ \hline
$((clk_{n},b_1),(clk_{n},b_2))$ & $\max(b_2*0-b_1*0,0)$\\
\hline\end{tabular}
\end{table}

Inspired by the formulation of AUC, at the first glance, we can define a \textbf{Real-Valued AUC} ($AUC^R$) by relaxing the $y_i$ in the original AUC from binary into real values.

\begin{equation}
\begin{aligned}
\label{func:5}
    AUC^R = \mathbb{E} [ & \mathbb{I}{(\widehat{y_i}>\widehat{y_j})} max(0,y_i-y_j) + \\
                                & \mathbb{I}{(\widehat{y_j}>\widehat{y_i})} max(0,y_j-y_i)]
\end{aligned}
\end{equation}

Note that the above definition is asymmetric as the correct ranking action being rewarded with $|y_i - y_j|$ while the incorrect one staying unpunished.
This asymmetry is degenerated in binary valued cases since the reward is either 1 or 0. 
However, it might be problematic in real valued cases.
To address this issue, the original {$AUC^R$} can be fixed as follows\footnote{If there is no special description, the following $AUC^R$ refers to this expression.}.

\begin{equation}
\begin{aligned}
\label{func:6}
    AUC^R = \mathbb{E} [ & \mathbb{I}{(\widehat{y_i}>\widehat{y_j})} (y_i-y_j) + \\
                                & \mathbb{I}{(\widehat{y_j}>\widehat{y_i})} (y_j-y_i) ]
\end{aligned}
\end{equation}

In sponsored search, AUC, especially AUC measured only on eCTR, may make some discrepancy and even produce misleading estimations when using it as the indicator for online RPM. Instead of characterizing click-through, $AUC^R$ depict the online RPM directly. The superiority of $AUC^R$ makes it more suitable for advertising scenarios and can be used as an offline measure of online RPM. The general solution of $AUC^R$ is described in \textbf{Algorithm 1}. The metric is bounded between $[-1,1]$ so that it can be used as an offline evaluation for online RPM.

\floatname{algorithm}{Algorithm}  
\renewcommand{\algorithmicrequire}{\textbf{Input:}}  
\renewcommand{\algorithmicensure}{\textbf{Output:}}  
  
    \begin{algorithm}  
        \caption{The algorithm of $AUC^R$}  
        \label{alg:1}
        \begin{algorithmic}[1] %每行显示行号  
        \Require A sequence of ($eCTR_i$,$click_i$,$bid_i$)
        \Ensure The offline evaluation metric $AUC^R$  
    \State Sort the sequence by $eRPM_i=eCTR_i*bid_i$ in descending order; $loss=0$
    \For{$i=0$; $i<m$; $i++$}
        \For{$j=i$; $j<m$; $j++$}
            \State $loss+=(click_j*bid_j-click_i*bid_i)$
        \EndFor
    \EndFor
       
    \State $AUC^R=\frac{1}{Z}loss$, where $Z=\widehat{loss}$ is the total loss of the sequence ranked by $RPM_i=click_i*bid_i$ in descending order.
    \State return $AUC^R$
        \end{algorithmic}
    \end{algorithm}  

\begin{comment}
In practice, we design a new metric named $gAUC^R$, which is the generalization of $AUC^R$. The $gAUC^R$ is a weighted average of $AUC^R$ calculated in the subset of samples group by impressions, users and so on. An impression based $gAUC^R$ is calculated as follows:

\begin{equation}
\begin{aligned}
gAUC^R&=\frac{\sum_{i=1}^{m}w_i*{AUC_i}^R}{\sum_{i=1}^{m}w_i} \\
&=\frac{\sum_{i=1}^{m}impression_i*{AUC_i}^R}{\sum_{i=1}^{m}impression_i}
\end{aligned}
\end{equation}

The $gAUC^R$ is practically proven to be more indicative in sponsored search settings, where \textbf{function (1)} is applied to rank candidate ads for each impression and model performance is mainly measured by how good the ranking list is, that is , an impression specific $AUC^R$. Hence, this methods can remove the impact of impression bias and measure more accurately the performance of the model over all impressions. With lots of experiments in our system, $gAUC^R$ metric is verified to be more stable and reliable than $AUC^R$.
\end{comment}

\subsubsection{Soft AUC}
The $SAUC$ could be defined by replacing the hard indicator function in the original AUC with soft ones (e.g. sigmoid). In this way, $SAUC$ is derivable with respect to the parameters.

\begin{equation}
\begin{aligned}
  \SoftAUC =  \mathbb{E} [&\SwapIndicSoftY{i}{j} \SwapGainSym{i}{j}  + \\
                      &\SwapIndicSoftY{j}{i} \SwapGainSym{j}{i} ] 
\end{aligned}
\end{equation}

And an empirical $SAUC$ by involving a predictor parameter $\theta$ on sample set $\mathbb{S}$ could be defined as follows with $Z$ as normalizer.

\begin{equation}
\begin{aligned}
  {\widehat{\SoftAUC}(\theta; \SampleSet)} = \frac{1}{Z} \sum\limits_{i=1}^{|{\mathbb{S}}|} \sum\limits_{j=i+1}^{|{\mathbb{S}}|} [ &\SwapIndicSoftF{i}{j} \SwapGainSym{i}{j}  + \\
                                                                                                      & \SwapIndicSoftF{j}{i} \SwapGainSym{j}{i} ] 
\end{aligned}
\end{equation}

\subsection{Optimal Ranking Function}
\subsubsection{Explicit Ranking Function} Based on previous research results \cite{lr-ctr1,lahaie-1}, we combine the calibration methods and price-squashed factor to  develop an explicit ranking function for revenue management. In practice, the explicit ranking function is designed as follows.
\begin{equation}
\begin{aligned}
    Ranksore &= f(eCTR, bid) \\
             &= (eCTR + \tau_d (eCTR))^\beta * bid \\
             &= (eCTR^{'})^\beta * bid
\end{aligned}
\end{equation}
where $\tau_d (eCTR)$ is a calibration factor ($eCTR$ is the predicted CTR, $\beta$ is the price-squashed factor to tune the weight between eCTR and bid, and $eCTR^{'}$ is the calibrated eCTRs.  And we use a piecewise linear function to cope with the complicated shapes in bias curves. In this way, the definition of the problem (\textbf{function 3}) can be further refined as follows,
\begin{equation}
\begin{aligned}
\label{fun:9}
f^* &= \arg \max_{f\in\digamma} Revenue \\
&=\arg \max_{f\in\digamma}{AUC^R} \\
&=\arg \max_{f\in\digamma}{SAUC}
\end{aligned}
\end{equation}
where $\digamma$ is the ranking function with the form of \textbf{function 9}. Two algorithms are presented to find the optimal parameters.

\textbf{Grid Search Method} Since the $AUC^R$ is not derivable with respect to the parameter, we can only use the grid search methods to solve this problem. The detailed algorithm is shown in \textbf{Algorithm 2}.

\floatname{algorithm}{Algorithm}  
\renewcommand{\algorithmicrequire}{\textbf{Input:}}  
\renewcommand{\algorithmicensure}{\textbf{Output:}}  
    \begin{algorithm}[h]  
        \caption{The Grid search method for revenue management}  
        \label{alg:2}
        \begin{algorithmic}[1] %每行显示行号  
        \Require The sequence \{($eCTR_1^{'}$, $bid_1$, $click_1$), ..., ($eCTR_m^{'}$, $bid_m$, $click_m$)\}
        \Ensure Optimal $\beta $
        \State Initialization: initialize $\beta$=0, $AUC^R$=0, $\beta_{min}$=0.1 and $\beta_{max}$=2.0
            \For{$j=\beta_{min}; j<=\beta_{max}; j+=0.02$}
                \State Calculate $currentAUC^R$ with Algorithm 1
                \If{$currentAUC^R > AUC^R$}
                    \State $\beta = j$
                    \State $AUC^R=currentAUC^R$
                \EndIf
            \EndFor
        \State return $\beta$
        \end{algorithmic}
    \end{algorithm}

\textbf{Gradient Descent Method} Different from $AUC^R$, the $\EmpSoftAUCExp$ is derivable with respect to $\theta$, it could be maximized with gradient based methods.
However, the computational complexity is unacceptable for industrial problems with hundreds of millions of samples.
To tackle this issue, we adopts mini-batched gradient method (\textbf{Algorithm 3}).
At the beginning, the whole sample set $\SampleSet$ is randomly divided into a series of sub sets, each of which contains tractable number(e.g. 100) of samples.
Then those sub sets are repeatedly fed into the optimizer and the $\theta$ is consistently updated until convergence. Experimental results show that our method converges to $\theta^*$ with both $AUC^R$ and $\SoftAUC$ maximized. 
%The detailed algorithm of this method is shown in \textbf{Algorithm 3}.

\floatname{algorithm}{Algorithm}  
\renewcommand{\algorithmicrequire}{\textbf{Input:}}  
\renewcommand{\algorithmicensure}{\textbf{Output:}}  
  
    \begin{algorithm} [h]  
        \caption{Mini-Batched Gradient Descent Method for revenue management }  
        \label{alg:1}
        \begin{algorithmic}[1] %每行显示行号  
        \Require sample set $\SampleSet$
        \Ensure $\theta^*$ which minimizes -$\EmpSoftAUCExp$ 
    
    \State Initialize $\theta$ randomly
    \State Split $\SampleSet$ into $K$ sub sets $\SampleSet_1$, ... , $\SampleSet_K$
    \While{not converged yet}
        \For{$k=1$; $k<=K$; $k++$}
            \State Compute $\Delta = -\frac{\partial{\EmpSoftAUC (\theta; \SampleSet_k)}}{\partial{\theta}}$
            \State Update $\theta = \theta - \eta \Delta$
        \EndFor
    \EndWhile

    \State return $\theta$
        \end{algorithmic}
    \end{algorithm}  

\subsubsection{Implicit Ranking Function} The popularity of deep learning has attracted the attention of countless researchers. One of the most impressive facts about neural networks is that they can fit to any function. That is, no matter what the function, there is guaranteed to be a neural network so that for every possible input, $x$, the value $f(x)$ (or some close approximation) is output from the network \cite{deep-network}. Given the power of deep networks, we explore to learn a reasonable ranking function with deep model. The structure of the model is show in \textbf{Figure 1}. In practice, we use the wide and deep networks \cite{deep-ctr3} to train a CTR prediction model. And the estimated CTR and the bid are fully connected with 3-hidden layers. The loss function of the task is the proposed $SAUC$ and we use AdaGrad to learn the implicit function of eCTR and bid.

\begin{figure}[ht]
\centering
\includegraphics[scale=0.40]{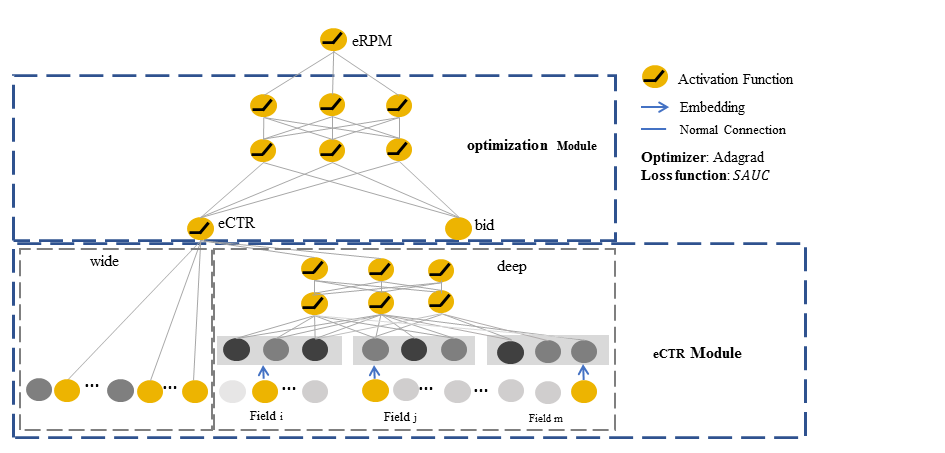}
\caption{The deep networks to learn an implicit ranking function}
\label{fig:label}
\end{figure}

\subsection{Summary of The Methods}
This paper has two major innovations: 1) From the view of loss function, we propose metrics for click that are based on final auction performance. The metrics are addressed to indicate the online RPM directly instead of the CTR, which is significant in sponsored search. 2)We explore the implicit and explicit ranking functions to maximize the online RPM. The summary of the methods is shown in \textbf{Table 3}.

\begin{table}[!htb]
\centering
\caption{Summary of the methods}
\label{tab:3}
\begin{tabular}{c|c|c|c} \hline
\textbf{Methods}&\textbf{Metric}&\textbf{Rank Func} & \textbf{Optimization}\\ \hline \hline
$M_1$& $AUC^R$& Explicit& Grid Search \\ 
$M_2$& $SAUC$& Explicit& Gradient Descent \\ 
$M_3$& $SAUC$& Implicit& Gradient Descent\\ 
\hline
\end{tabular}
\end{table}

\section{Experiments and benchmarking}
We depict our experiments for benchmarking the proposed methods in this section. The experimental results indicate that our proposed metrics are more effective than any other state-of-the-art metrics. Further, we make an exploration and discussion for the optimal ranking function.

\subsection{Experimental Data Set}
Throughout the paper we show motivating examples and the analyses of the model performance on two e-commerce search engine, www.alibaba.com and www.aliexpress.com\footnote{We intend to make the data and code available for open research.}.
In particular, www.alibaba.com is a b2b (business to business) e-commerce search engine and www.aliexpress.com is mainly about b2c (business to customer). The experimental results on the two cross-domain platforms demonstrate the generality of the presented methods.

\subsection{Performance of the Evaluation Metrics}
In this section, we evaluate the proposed metrics and existing metrics (mainly AUC) on the two data sets listed in Section 4.1, and compare their performance. We use a \textbf{Confusion Matrix} to measure the performance of the offline evaluation metric (\textbf{Table 4}). The matrix can be very intuitive to show the performance of online and offline. Our platforms have hundreds of pages and traffic sources and we select the most important 50 pages from Aliexpress platform and 30 pages from Alibaba platform as the experimental environment. The selected pages contribute to the main revenue of the platforms (91.5\% and 70.6\% respectively), and their traffic distributions are relatively stable, which is suitable for comparison experiments. We train a new model and tune parameter settings based on historic logs data to collect a set of experimental data. In order to verify the stability and effectiveness of the proposed metrics, we collect experimental data from the productive system for 14 days. We finally collect 1120 sets of experimental data. \textbf{Table 4} summarize offline and online matrices which are tested on online A/B testing environments on Alibaba and Aliexpress with real-time user traffic.

\begin{comment}
A serious problem with $AUC$ in practice is the discrepancy in performance between the offline and online testing. There are cases where a predictive model that achieved significant gain on offline metrics does not perform as well or sometimes even under-perform when deployed on the online system.
\end{comment}

\begin{table}[!htb]
\caption{The Confusion Matrices of different evaluation metrics}  
\centering  
\begin{threeparttable}
\subtable[AUC Matrix]{  
       \begin{tabular}{p{2.2cm}<{\centering}|p{2.2cm}<{\centering}|p{2.2cm}<{\centering}} \hline
       \label{tab:4:a}
       & \textbf{Online+}\tnote{1} & \textbf{Online-}\tnote{2}\\ \hline
        \textbf{Offline+}\tnote{3} & 504 & 89 \\ \hline
        \textbf{Offline-}\tnote{4} & 22&  505\\ \hline
        \end{tabular} 
}  
\qquad  
\subtable[$AUC^R$ Matrix]{          
       \begin{tabular}{p{2.2cm}<{\centering}|p{2.2cm}<{\centering}|p{2.2cm}<{\centering}} \hline
       \label{tab:4:b}
       & \textbf{Online+} & \textbf{Online-}\\ \hline
        \textbf{Offline+} & 581 & 12 \\ \hline
        \textbf{Offline-} & 4 & 523 \\ \hline
        \end{tabular} 
}  
\qquad  
\subtable[$SAUC$ Matrix]{          
       \begin{tabular}{p{2.2cm}<{\centering}|p{2.2cm}<{\centering}|p{2.2cm}<{\centering}} \hline
       \label{tab:4:c}
        & \textbf{Online+} & \textbf{Online-}\\ \hline 
        \textbf{Offline+} & 582 & 11 \\ \hline
        \textbf{Offline-} & 4 & 523 \\ \hline
        \end{tabular} 
}  
    \begin{tablenotes}
        \footnotesize
        \item[1] The online performance (RPM) is better than baseline.
        \item[2] The online performance cannot beat baseline.
        \item[3] The offline performance (AUC, $AUC^R$ and $SAUC$ as the metrics respectively) is better than baseline.
        \item[4] The offline performance cannot beat baseline.
      \end{tablenotes}
\end{threeparttable}
\end{table}

\begin{comment}
From \textbf{Table \ref{tab:4:a}} we can draw the conclusion that the $RPMLOSS$ and $bRPMLOSS$ are more effective metrics to measure online performance in sponsored search. The possible reasons are as follows. Firstly, the LOG-LOSS just depicts the loss of clicking or not clicking which may not be suitable in sponsored search when considering the impact of bidding. Secondly, we use observed-RPM instead of tRPM when calculating $RPMLOSS$ and $bRPMLOSS$, which can effectively reduce sparsity of the object function. 

From \textbf{Table \ref{tab:4:a}}, we can also observe that $bRPMLOSS$ has better performance than $RPMLOSS$. Through in-depth analysis, we give the follow explanation. We use "bidprice" and "clickprice" when calculating $RPMLOSS$ and $bRPMLOSS$ respectively. The "clickprice" is the actual benefit obtained by our platform and $bRPMLOSS$ is more realistic about the online situation. In other words, $bRPMLOSS$ has a better characterization of the real benefits of the advertising system. The conclusion is further proved in \textbf{Table \ref{tab:4:b}} ($bRPM_{auc}$ and $gbRPM_{auc}$ are much better than $RPM_{auc}$).
\end{comment}

As \textbf{Table 4} shows, while AUC is a quite reliable method to assess the performance of predictive models, it still suffers from drawbacks in the sponsored search. From the table, we can draw the following conclusions:
\begin{enumerate}
    \item It has been observed that higher AUC dose not necessarily mean better ranking always. In our observation, 8\% ($89/(504+89+22+505)$) of the samples perform well in the offline but bad on the online system. 
    \item On the other hand, a lower AUC dose not necessarily mean a worse performance in the online environment. 2\% ($22/(504+89+22+505)$) of the inconsistencies may be quite misleading, and we may miss the optimal solution (In practice, AUC as the indicator, the 2\% will not be released online). Despite the widespread use, in general, the AUC is neither sufficient nor necessary for online performance in sponsored search.
    \item The proposed metrics eliminate the discrepancy between online and offline greatly. The $AUC^R$ and $SAUC$ have a comparable performance, and beat the AUC well. The main reason is that $AUC^R$ and $SAUC$ try to model the performance of the whole ranking function, while AUC merely measures the accuracy of the click-though rates without considering the factor of the bid.
\end{enumerate}

\subsection{Convergence of $SAUC$}

\begin{figure*}[h]
\centering
\includegraphics[scale=0.40]{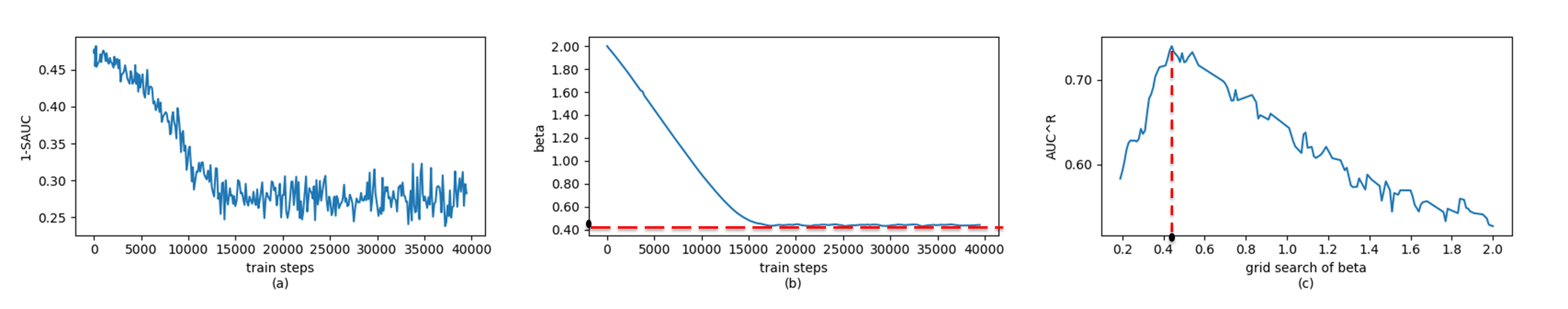}
\caption{Convergence of $SAUC$ and the equivalent of \textbf{Algorithm 2} and \textbf{Algorithm 3}}
\end{figure*}

\begin{table*}[h]
\centering
\caption{Performance of A/B test with real-world traffic}
\begin{tabular}{c|c|c|c|c|c|c|c|c} \hline
 & & &\multicolumn{3}{c|}{\textbf{Alibaba.com}} & \multicolumn{3}{c}{\textbf{Aliexpress.com}} \\ \hline 
 \textbf{Methods} & \textbf{Rank function} & \textbf{Objective Metric} & \textbf{RPM} & \textbf{CTR}& \textbf{CPC} & \textbf{RPM} & \textbf{CTR} & \textbf{CPC} \\ \hline 
 \textbf{Baseline} & $eCTR^\beta*bid$ & Manual tuning &- &- & -&- &- &- \\ \hline
 \textbf{Method1} & $eCTR^\beta*bid$ & $SAUC$& +9.65\%&+11.68\% &-2.14\% &+10.01\%  &+28.48\% &-14.98\% \\ \hline
 \textbf{Method2} & $(eCTR^{'})^{\beta}*bid$ & $SAUC$ & +9.92\% & +12.71\% &-3.68\% & +12.97\% & +31.97\% & -14.60\%  \\ \hline
\end{tabular}
\end{table*}

The $SAUC$ is proposed to solve the non-derivable issue of $AUC^R$, so that we can use gradient descent method to optimize the parameters. Theoretically, the $SAUC$ is approximately equivalent to $AUC^R$. In this paper, we design an experiment to verify the convergence of $SAUC$ and the equivalence of $SAUC$ and $AUC^R$. In the experiment, the ranking function is the form of \textbf{function 2}. And we use \textbf{Algorithm 2} and \textbf{Algorithm 3} to optimize the parameter respectively. The experimental result is shown in \textbf{Figure 2}. Comparing \textbf{Figure 2(a)} and \textbf{Figure 2(b)} we can conclude that $SAUC$ converges well. Comparing \textbf{Figure 2(b)} and \textbf{Figure 2(c)} we come to further conclusion that \textbf{Algorithm 2} and \textbf{Algorithm 3} have the equivalent performance, and the optimal value of $\beta$ is 0.43.

\subsection{Exploration of Ranking Function}
Motivated by \cite{lahaie-1,lahaie-2,rank-function2}, we design an explicit ranking function (\textbf{Function 9}). 
%In our method, the solution process is divided into two stages. We use the spline regression to learn the $eCTR^'$, and then optimize the parameter $\beta$ with the proposed algorithms (\textbf{Algorithm 2} and \textbf{Algorithm 3}). 
In view of the power of deep network, we design a deep model to learn the optimal implicit ranking function. By comparing these two methods, we come to some interesting findings which are shown in \textbf{Figure 3}. From the figure, we can draw the following conclusions:
\begin{enumerate}
    \item Theoretically, the explicit ranking function is a special case of the implicit ranking function. However, experimental results show that the designed ranking function and the model-based function have a considerable performance. The two approaches converge to the same optimal value.
    \item The implicit ranking function convergence faster than the explicit one. Multiple rounds of experiments show that deep networks make it easier to capture the functional relationship between $eCTR$ and $bid$.
\end{enumerate}

\begin{figure}[h]
\centering
\includegraphics[scale=0.28]{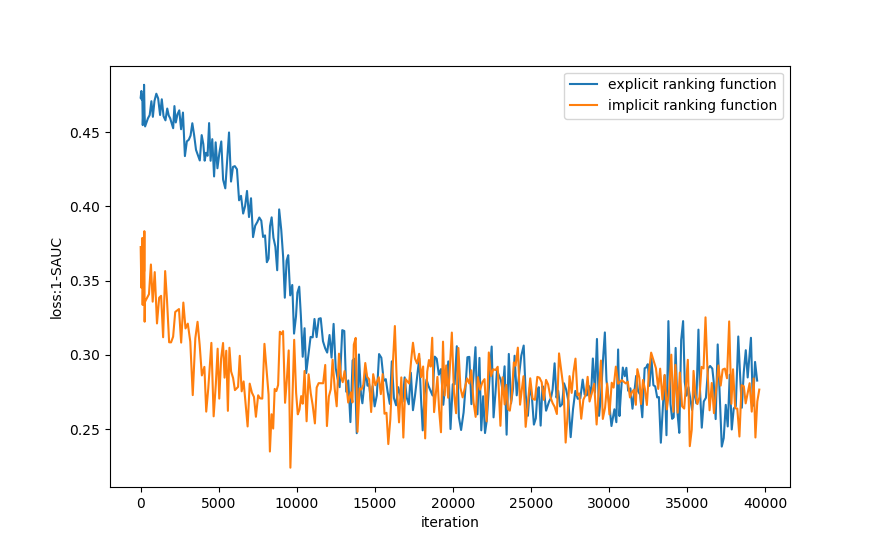}
\caption{Exploration of Ranking Function}
\end{figure}

\begin{comment}
\begin{figure*}
\centering
\subfigure[Alibaba.com]{
\label{fig:2:a} %% label for first subfigure
\includegraphics[width=16cm]{sc_full.png}}
\hspace{1in}
\subfigure[AliExpress.com]{
\label{fig:2:b} %% label for second subfigure
\includegraphics[width=16cm]{ae_full.png}}
\caption{The experimental results of A/B test with real-world traffic}
\label{fig:2} %% label for entire figure
\end{figure*}
\end{comment}

\subsection{Performance of Online A/B Test}
We have deployed our proposed strategies on Alibaba and AliExpress platforms, which are two mainstream platforms in the global e-commerce market. 
%Both training and testing data sets are generated from system logs, including impression, click and bidding logs. 
The $AUC^R$ and $SAUC$ are used as the final evaluation metrics in our system. In view of the performance,  we use the explicit ranking function in our production system finally. The proposed approach can be regarded as a post-processing process based on the existing click-through prediction model. For the sake of comparability, the baseline models and our proposed model are constructed on the same feature representation. Parameters are tuned separately and we report the best results.

\begin{comment}
\begin{table}[h]
\centering
\caption{Performance of A/B test with real-world traffic}
\label{tab:1}
\begin{tabular}{c|c|c|c} \hline
\textbf{Platform}&\textbf{RPM} &\textbf{CTR}&\textbf{CPC}\\ \hline \hline 
Alibaba.com   &  +9.92\% & +12.72\% & -3.68\%\\ \hline
Aliexpress.com  &  +12.97\% & +31.97\% &-14.60\%     \\

\hline\end{tabular}
\end{table}
\end{comment}

\textbf{Table 5} is the A/B test result of online systems. We show performance over three metrics. The experimental results show that the methods described in this paper outperform the state-of-the-art models on RPM, CTR and CPC. The proposed methods have brought significant improvement to the RPM of the platforms. It is well to be reminded, our direct optimization object is the platform's RPM. From the results, however, we can illustrate that we achieved the goal without compromising the advertiser's benefit and the customer's search experience. On the contrary, we improved the CTR and advertisers' ROI at the same time. 
%We obtained the three-way profitability of platform's revenue, advertiser's effect and buyer's search experience.

\section{Conclusion and discussion}
In this work we looked into the revenue management problem that contains the Alibaba and Aliexpress as special cases. From the view of loss function, we propose two metrics, $AUC^R$ and $SAUC$ for click modeling that are based on final auction performance. The metrics are potentially more optimal than AUC because the goal is to depict the online RPM directly. A lot of theoretical analysis and experimental results verify the superiority of the proposed metrics as an indicator for the online RPM. We also explored the ranking functions, both implicit and explicit ones, to maximize the revenue in sponsored search. The methods are deployed on two production platforms. Outstanding profit gain over the baseline were observed in online A/B tests with real-world traffic. 
%On the other hand, experimental results show that we improve the revenue ability of the platform without compromising the advertiser's benefit and the customer's search experience.

For future work, we will analyze the factor of position bias \cite{position-2,position-1} in modeling the revenue management. We also plan to further explore the implicit ranking functions to maximize the online revenue.

%ACKNOWLEDGMENTS are optional
\begin{comment}
\section{Acknowledgments}
This section is optional; it is a location for you
to acknowledge grants, funding, editing assistance and
what have you.  In the present case, for example, the
authors would like to thank Gerald Murray of ACM for
his help in codifying this \textit{Author's Guide}
and the \textbf{.cls} and \textbf{.tex} files that it describes.
\end{comment}

\bibliographystyle{named}
\bibliography{sigproc}  % sigproc.bib is the name of the Bibliography in this case
% ACM needs 'a single self-contained file'!
%\balancecolumns
%\appendix
%Appendix A
%\section{Headings in Appendices}
%\balancecolumns % GM June 2007
% That's all folks!

\end{document}